\documentclass[11pt,a4]{article}
\usepackage[hyperref]{acl2021}
\usepackage{times}
\usepackage{latexsym}
\usepackage{dblfloatfix}
\usepackage{graphicx}
\usepackage[T1]{fontenc}
\usepackage{csquotes}
\usepackage{xcolor}
\usepackage{soul}

\usepackage{microtype}
\aclfinalcopy
\MakeOuterQuote{"}

\title{Using Machine Translation to Localize Task Oriented NLG Output}

\author{Scott Roy  \quad {\bf Cliff Brunk} \quad {\bf Kyu-Young Kim } \\ {\bf Justin Zhao} \quad {\bf Markus Freitag} \quad {\bf Mihir Kale} \quad {\bf Gagan Bansal} \quad {\bf Sidharth Mudgal} \quad {\bf Chris Varano} \\\\
Google, Inc.}

\date{}

\begin{document}
\maketitle
\pagestyle{plain}

\begin{abstract}
One of the challenges in a task oriented natural language application like the Google Assistant, Siri, or Alexa is to localize the output to many languages. This paper explores doing this by applying machine translation to the English output. Using machine translation is very scalable, as it can work with any English output and can handle dynamic text, but otherwise the problem is a poor fit. The required quality bar is close to perfection, the range of sentences is extremely narrow, and the sentences are often very different than the ones in the machine translation training data. This combination of requirements is novel in the field of domain adaptation for machine translation. We are able to reach the required quality bar by building on existing ideas and adding new ones: finetuning on in-domain translations, adding sentences from the Web, adding semantic annotations, and using automatic error detection. The paper shares our approach and results, together with a distillation model to serve the translation models at scale.

\end{abstract}

\section{Problem Overview}
The problem this paper explores is to localize the output of a task oriented natural language application like the Google Assistant, Siri, or Alexa. These are examples of very broad task oriented systems, but there are also much narrower ones, for example an application for movie reservations. We refer to applications like these as task oriented NLG applications, since one of the main requirements is to generate natural language text. In task oriented NLG, the user speaks or types in natural language, and the system responds in kind. A typical architecture is to use a pipeline of components to go from speech, to text, to a semantic interpretation on the input side, and from structured data, to text, to speech on the output side. \cite{kukich1983design,mckeown1985text} This paper focuses on the data to text part of this pipeline, with the specific goal of figuring out how to automatically produce text in many different languages. The approach we take is to use neural machine translation, NMT \cite{sutskever2014sequence}, using a variety of techniques to reach a very high quality bar.

Data-to-text is a form of natural language generation where the input is structured data like a table or a graph. We assume that we have a system that can produce English output, and that the input to our localization problem is a combination of structured data, English text, and a target language. The output should be a fluent, grammatical, natural language sentence in the target language.

In our particular task the structured data is a list of key/value arguments. For example:

\begin{description}
\item[Structured data] \hfill \\
	intent: CURRENT\_TEMP \\
	date: "20190717" \\
	temperature: 25 \\
	temp\_unit: CELSIUS \\
	humidity: 83 \\
	current\_date: "20190717" \\
	location\_en: "Caymen Islands" \\
	location\_de: "Caymen-Inseln"
\item[English] \hfill \\
    "It is currently 25 degrees in the Cayman Islands."
\item[Target language] \hfill \\
    de
\item[Target text] \hfill \\
    "Es ist gerade 25 Grad auf den Caymen-Inseln."
\end{description}

\begin{figure*}
\includegraphics*[page=1,trim=0in 1.75in 0in 1.25in, width=6in]{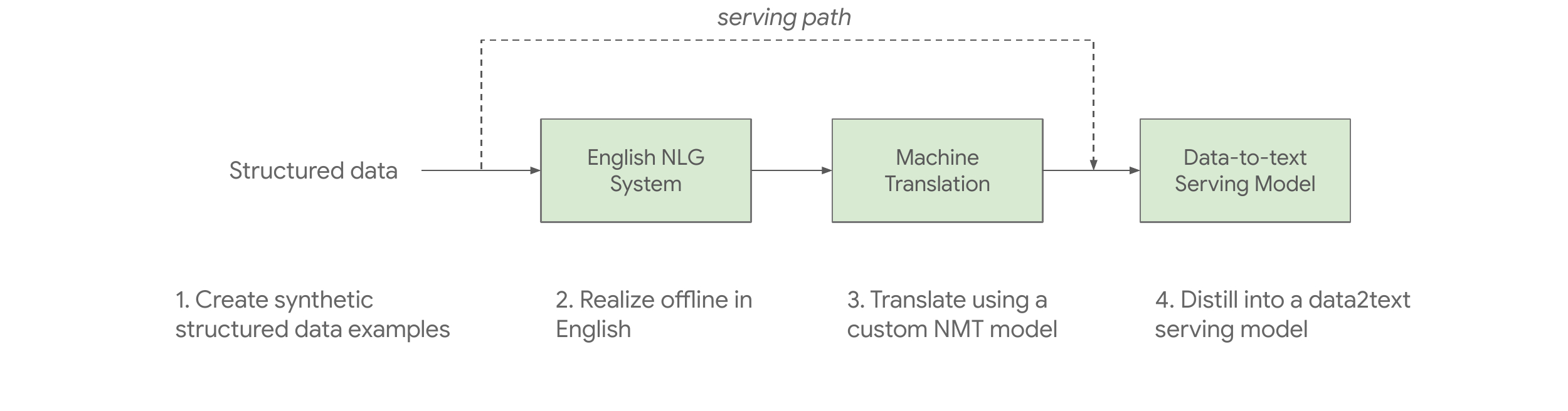}
\caption{Training and inference approach}
\label{architecture}
\end{figure*}

\noindent The arguments can be strings, numbers, or enum values. The schema of possible arguments is fixed, but different examples can include different arguments. Also, the natural language output may not use all of the arguments, and some arguments may be redundant. We expect to have both English and localized text for string arguments.

The core of our localization approach is to train a custom NMT model that works on the English output, and that we finetune to work well for the sentences we need to translate. This model is very high quality, but slow to serve directly, so we use it offline to create training targets for a distilled model \cite{hinton2015distilling} that goes from the structured data directly to localized text. The complete training architecture is shown in Figure~\ref{architecture}.

While our approach works for any language, we present aggregate results for 8 low resource Indic languages that we have focused on: Marathi, Bengali, Tamil, Telugu, Gujarati, Kannada, Malayalam, and Urdu. These languages are challenging because the baseline NMT quality is low \cite{siripragrada2020multilingual}, and because it is difficult to find text and linguistic resources for them.

Machine translation is actually a poor fit for our problem. Even though the quality of NMT has improved over the years \cite{chen2018best}, general purpose NMT models tend to perform poorly in the presence of domain mismatch \cite{koehn2017six}. Based on our experience, the domain of task oriented responses is indeed very different than the typical data used to train an NMT system. Moreover, for a task oriented NLG application the quality often needs to be nearly perfect in every way: grammatical, fluent, and accurate. This is an extremely high bar to reach even for head languages where machine translation works well, let alone for the low resource languages that we focus on.

\begin{figure}[h]
\includegraphics*[page=2,width=3in,trim=0.5in 1in 0.5in 0in]{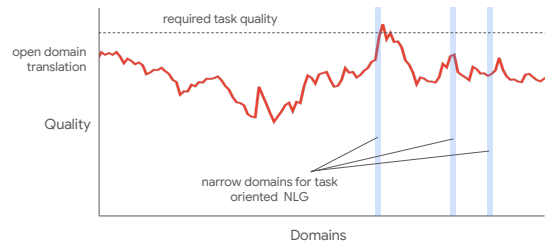}
\caption{Coverage and quality for task oriented NLG}
\label{narrow-range}
\end{figure}

Fortunately, in a task oriented NLG application, the range of sentences that we need to translate is typically very narrow. We usually only care about a small number of domains, for example restaurant reservations or reporting the weather, and a limited number of intents in each domain, for example giving the current temperature or tomorrow's forecast. The overall situation is therefore like Figure~\ref{narrow-range}. This shows the contrast between the enormous range of sentences that we need to handle for open domain machine translation, and the very narrow range we need to handle for the domains in a task oriented NLG application. It also shows how NMT quality varies greatly by domain. The figure shows three domains, but even if we assume there are dozens, and that there are a few dozen distinct intents for each domain, and a few distinct sentence variations for each intent, there are still only O(10k) sentences if we ignore the differences in the argument values and grammatical inflection. Our goal is to improve the quality for this narrow range of sentences to meet a quality bar that is typically much higher than we can achieve with general purpose machine translation.

Our approach builds on existing domain adaptation ideas in machine translation \cite{freitag2016fast, chu2018survey}, refining, combining, and developing new ideas to meet the unique requirements of our problem. The main contributions of the paper are:

\begin{itemize}
    \item Showing how to combine domain adaptation and NMT quality ideas like back translation, finetuning on in domain sentences, and semantic annotations to reach the quality needed for a task oriented NLG application.
    \item A new accuracy error detection model.
    \item An attention based data-to-text model that directly encodes the structure in the input.
\end{itemize}

\section{Improving NMT Quality}
There are many kinds of errors we need to fix to reach the desired quality. After looking at the NMT errors flagged by human raters on several thousand sentences in our task, we have focused on four main kinds. We illustrate with examples in English so that it's easy to understand the errors. In all cases the correct sentence we are trying to produce is "It is 72 and sunny in Mountain View today.":

\definecolor{light-red}{RGB}{244,204,204}
\sethlcolor{light-red}

\begin{description}
\item[Grammar mistakes] \hfill \\
	“It \hl{are} 72 and sunny in Mountain View today.”
\item[Poor word and phrase choices] \hfill \\
	“It will be 72 and \hl{no precipitation in the atmosphere} in Mountain View today.”
\item[Literal entity translations] \hfill \\
	“It will be 72 and sunny in \hl{Pinnacle Vista} today.”
\item[Accuracy errors] \hfill \\
	“It will be \hl{27} and \hl{rainy} in Mountain View today.”
\end{description}

\noindent We've come up with different ways to fix each of these kinds of errors.

\subsection{Fixing grammar mistakes}
\label{nmt-grammar}
To fix grammar mistakes, we include a large number of high quality sentences from the Web in the target language. The idea is that if the NMT model sees enough grammatical sentences in the target language, it will develop a strong language model that makes it hard for it to produce ungrammatical sentences.

We add the Web sentences to the NMT training data using back-translation \cite{sennrich2015improving}. We run each sentence through an existing off-the-shelf NMT model to translate it into English, and then include the resulting sentence pairs in the training data for our custom model. The backtranslation NMT model has the same architecture as the custom one that we describe in section~\ref{nmt-arch}.

We find sentences to include by running a custom sentence extractor on Web pages, using a language ID model to filter for the target languages. We then compute a quality score for each sentence. The quality score comes from a Bayesian classifier that compares each sentence to sentences in a hand annotated set of high quality, in-domain Web pages using unigram and bigram word features. The quality score is the log odds ratio that a sentence comes from the in-domain set rather than from the larger pool of all Web sentences:
\[
    score = \sum_{w \in S}log\left(\frac{p_D(w)}{1-p_D(w)}\frac{1-p_C(w)}{p_C(w)}\right)
\]
\noindent The sum is over all of the unigram and bigram words $w$ in a sentence $S$, and $p_D(w)$ and $p_C(w)$ are the probability of seeing the word in the in-domain set and in the corpus as a whole.

For our NLG application the sentences we need to translate look like high quality sentences from news articles, blog posts, Wikipedia pages, and the like, so these are the sentences that we use for the in-domain set in the Bayesian classifier. However, for a different application we could use a different set of sentences, and this would help to create a language model more appropriate to that application.

Figure~\ref{sample-sentences} shows example sentences in English with their quality scores. We filter the sentences by discarding anything with a negative score. Although we have not done a formal evaluation, our experience is that this filtering does a good job of eliminating low quality, ungrammatical sentences that we do not want to use. The filtering eliminates about half of the sentences, leaving us on average with about 500M sentences in each of our 8 languages.

\begin{figure}[t]
\includegraphics*[page=3,width=3in,trim=2in 0.5in 2in 0in]{figures.pdf}
\caption{Sample Web sentences with quality scores}
\label{sample-sentences}
\end{figure}

\subsection{Fixing poor word and phrase choices}
We fix poor word and phrase choices by collecting new human translations of sample sentences in our NLG application. The great majority of bad translations come from application specific words and phrases, and so when we include good translations of sample sentences the model learns the right way to translate these for our task.

To collect the new translations, we sample one sentence for each domain and intent in our NLG application. As we noted earlier, even in a large task oriented NLG application there are only about 10,000 sentence variants, so this is a manageable amount of data. We use logs and a synthetic example generator that we describe in section~\ref{synthetic-examples} to create a structured example for each intent, and then run the examples through our English data-to-text system. This gives us one new sentence for each intent. The examples use diverse argument values to ensure that the resulting sentences are diverse, and so that when we train the NMT model it focuses on the common words and phrases in the NLG task rather than the argument values. We send all of the resulting sentences to human translators.

We finetune our NMT model on the new translations using the technique of contrastive data selection, CDS \cite{wang-etal-2018-denoising}. This lets us magnify a small number of high quality human translations by finding similar sentences in the far larger pool of Web sentences.

CDS works by starting with our base translation model X and doing a single finetuning step on the new human translations. This gives us a new model Y. We then score all of the original NMT training data by comparing the original probability of each target sentence using X with the new probability using Y. The sentences where the probability goes up the most are presumably the most like our new high quality translations, and so these are the ones that we weigh the most for finetuning. We now run a full finetuning on model X using the computed CDS weights to get our final NMT model.

\subsection{Fixing literal entity translations}
We see many errors where our baseline NMT model fails to recognize that a span of text is an entity, and so translates it literally. This is an acute problem in a task that contains many entities. If we have additional context that tells us which spans are entities, we can fix the problem by marking the spans:

\begin{quote}
    It's 72 and sunny in \textcolor{blue}{{\textless}location{\textgreater}}Palo Alto\textcolor{blue}{{\textless}/location{\textgreater}} today.
\end{quote}

\noindent In a task oriented NLG application, we typically know which spans of text are entities because of the structured data. We can mark a span as an entity if it exactly matches an argument that is an entity, or if we are able to get this information from the English data-to-text system, or by using a named entity recognition (NER) system. In our task, we get the entity spans at inference time from the English system, and we use an existing NER system at training time to label the entity spans in the English part of the NMT training data. The NER system is trained using the method from \citet{tsai-etal-2019-small} and fine-tuned with labeled entity mentions in 15 languages. We run the NER system in a high precision mode, since the labels that we get at inference time from the English system are high precision.

\subsection{Fixing accuracy errors}
We can typically trace accuracy errors at inference time to accuracy errors in the NMT training data. To fix these errors, we train a classifier to find the accuracy errors in the NMT data, then use it to filter out the bad pairs.

\begin{figure}[h]
\includegraphics*[page=4,width=3in,trim=2.25in 1.5in 2.25in 0.25in]{figures.pdf}
\caption{Training data for accuracy error model}
\label{accuracy-errors}
\end{figure}

The classifier takes in a translation pair and outputs the probability that the pair is an accurate translation. We train it by providing positive and negative examples as shown in Figure~\ref{accuracy-errors}. The positive examples are existing translation pairs, which we assume are accurate. The negative examples are these same pairs where we change a word in the English side to artificially create an accuracy error. Even if some of the positive examples are actually inaccurate, the negative examples have very little noise because of the way we construct them, and we have found that the final model works well despite the potential noise in the positive examples.

\begin{figure}
\includegraphics*[width=\columnwidth]{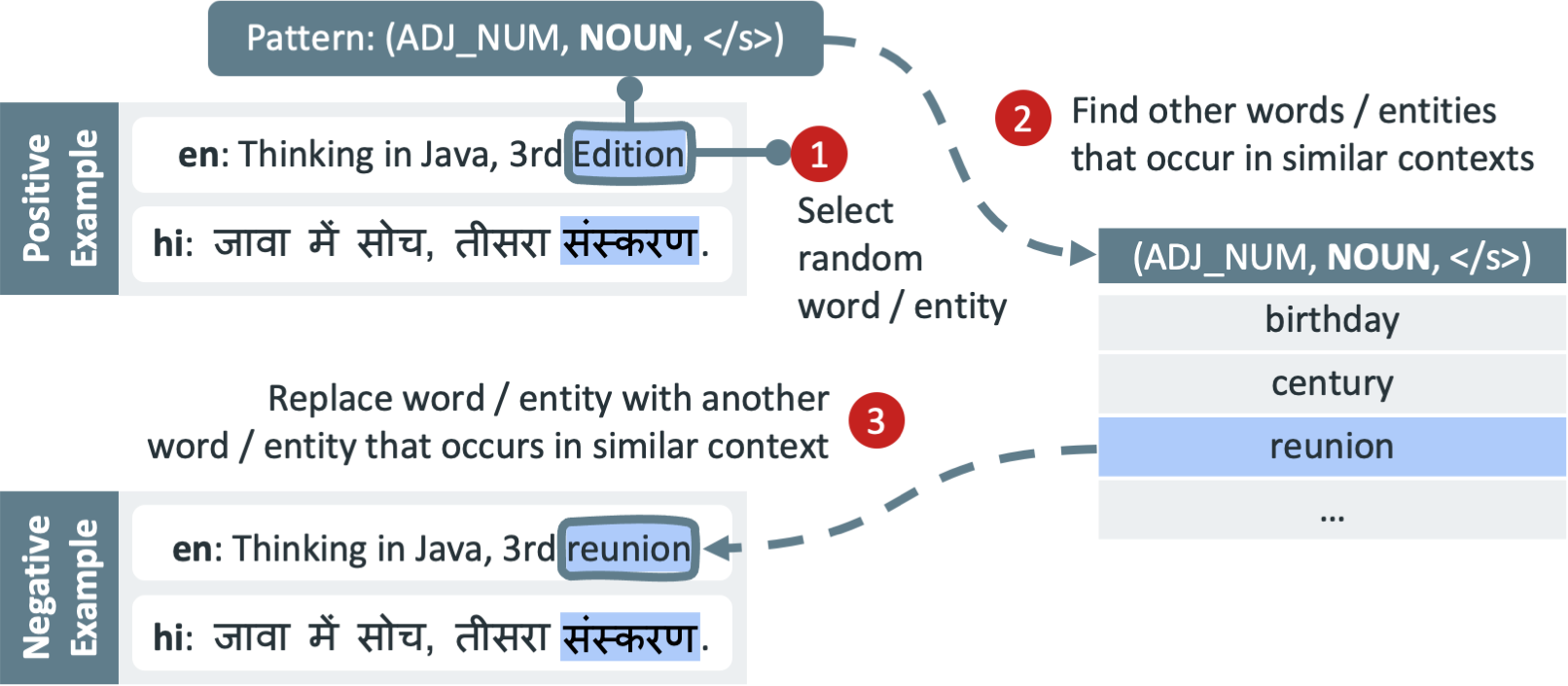}
\caption{Creating accuracy errors using POS and entity trigrams}
\label{trigram-replacement}
\end{figure}

We create the actual training examples in two different ways. The first uses the training data for our NMT model. To create the negative examples for this data, we use a POS and entity type tagger to annotate the words and spans in the English text. We then pick a word or phrase that is not a function word and randomly replace it with a word that appears in the same trigram context somewhere else in the NMT data, as shown in Figure~\ref{trigram-replacement}. This results in a new English sentence that is likely to be fluent but that modifies one of the content words, so the original translation is inaccurate. With this technique, we are able to generate 100-200M training examples per language.

The second way we create training examples is to use our NMT model. We create synthetic examples for our NLG application, run them through the English system, and translate them with our NMT model to get translation pairs. To create the negative examples for this data, we find two examples where the English sentences have the same grammatical form but different argument values. We then swap their translations. Using this technique, we are able to generate about 1-5M training examples per language.

The classifier is a text-to-text model where the input is "<English> SEP <Translation>", and the target is "CORRECT" or "INCORRECT". Empirically we find that training a model from scratch with this representation is quite unstable since the model has to learn all the intricacies of language from just the binary labels. Motivated by the findings in \citet{kale-roy-2020-machine}, we instead start with a target-language-to-en NMT model that already understands a great deal about both English and the target language. We then finetune the model on the training data described above. This NMT model has the same architecture and training data as the one described in the next section.

We use two different data sets to evaluate the model. The first is from our NLG application. We collected human translations of English task output in Bengali, Gujarati, Tamil, and Malyalam, on average 1300 sentences in each language. 27\% of these sentences contain accuracy errors, as we asked the translators to change the English entities to ones that would be common in-locale. Our model has 100\% recall finding these accuracy errors, while only incorrectly labeling 3.4\% of the correct translations as errors. The second data set is from the Web. We collected 5000 sentences and translated them into each of our 8 Indic languages, preserving the English entities so that there are no known accuracy errors. Our model incorrectly labels 10.2\% of these translations as errors. Looking at the errors, most of the mistakes are from idiomatic translations. Our takeaway from these results is that our model has very high recall and somewhat lower precision. This is what we want, since our goal is to filter out bad NMT training examples. It's okay to remove good ones too as long as there is no systematic bias to what we remove.

Anecdotally, we have also seen the model perform well on examples with hallucinations and omissions, where we either add information that is not in the English text or leave something out. It is good but occasionally misses number errors and copying mistakes. We have not formally evaluated these other kinds of errors.

\section{Machine Translation Model}
\subsection{Architecture and training methodology}
\label{nmt-arch}
We have trained an NMT model for each of our 8 Indic languages using the techniques in the previous sections. The model is a 6 layer Transformer Big encoder with an 8 layer RNMT+ decoder as described in \citet{chen-etal-2018-best}. For each language, we use approximately 200-300M parallel sentences, where about 10M are natural parallel data and the rest are our back-translated Web sentences. The fine tuning data consists of 1-5k high quality in-domain sentences.

\begin{table*}
\includegraphics*[page=6,width=6in,trim=1in 3in 1in 1in]{figures.pdf}
\caption{Evaluation results for the task specific NMT model}
\label{nmt-results}
\end{table*}

\subsection{Experimental results}
We have evaluated our model two different ways: using BLEU score comparisons against human translations, and human evaluation.

\subsubsection{BLEU evaluation}
For the BLEU score comparison, we finetuned our NMT model using 1.5k in-domain human translations in each of our 8 Indic languages, and we tested on a synthetic heldout set of about 900 translations. We create synthetic examples by randomly sampling argument values and running them through our English NLG system, as we describe later. The test set uses heldout argument values that are not in the finetuning data, and it also includes sentences with unseen intents and domains so that we can test the generalization performance when we have no example sentences in the training data. The sentences are split about equally between existing intents, unseen intents in existing domains, and completely unseen domains.

The aggregate BLEU score results are in the left half of Table~\ref{nmt-results}. We see substantial gains in all of the metrics. The custom NMT model shows a nice ability to generalize to new intents and domains, preserving large gains against the baseline NMT model.

\subsubsection{Human evaluation}
For the human evaluation we used all of our human translations to finetune the NMT model, and we created a new test set of 600 synthetic examples. This test set also uses unseen values. About 10\% of the examples are for existing intents in the training data, about 10\% are for completely unseen domains, and the remaining 80\% are for unseen intents in existing domains.
 
Human evaluation in prior data-to-text work \cite{ferreira20202020, wen2015semantically} evaluates sentences along different dimensions like grammatically, fluency, correctness, and coverage. We follow this approach and ask raters to evaluate the generated text with respect to grammar, naturalness and accuracy, similar to \citet{shimorina2018webnlg}. The grammar and naturalness metrics are on a 5 point scale, where grammar looks at basic grammatical agreement and naturalness evaluates how similar a sentence is to one that a native speaker would use. Accuracy is a Boolean yes/no measure of whether a sentence is an accurate translation of the English. The human raters are all fluent speakers of the target languages.

The aggregate results are in the right half of Table~\ref{nmt-results}. The 68\% baseline NMT accuracy number is an estimate based on additional golden analysis of the Marathi and Malayalam output, as we found that the raters were confused by the accuracy instructions in this particular evaluation. We see substantial gains in all of the scores.

\begin{table}[b]
\includegraphics*[page=7,width=3in,trim=2.5in 2.75in 2.5in 1in]{figures.pdf}
\caption{Evaluation results vs. human translation}
\label{nmt-human-results}
\end{table}

In order to calibrate the grammar, naturalness, and accuracy scores against human performance, we did a third evaluation for two of our languages, Marathi and Malayalam. We evaluated a new set of 900 examples, including 300 real examples from logs, and a second set of 200 human translations of real examples. Table~\ref{nmt-human-results} shows the results. The Malayalam NMT scores are comparable to the human translations, and---somewhat strangely---the Marathi NMT scores appear even better than the human translations. This is odd, so we looked at the human Marathi translations more carefully and found that many of the sentences with low grammar scores have mixed language text, with entities in foreign languages. This confuses both the translators and the raters, so that the real grammar and naturalness scores should be somewhat higher. Based on the evaluation scores and inspecting all of the examples with low scores, our overall takeaway is that the NMT output for our task is comparable to human translation within the limits that our current evaluation process can measure. 

\section{Data-to-text Model}
To decrease the model size and latency, we distill the NMT model into a custom data-to-text model that goes directly from structured data to localized text. The data-to-text model is an RNN decoder that reads from the structured data at every time step, as illustrated in Figure~\ref{data-to-text-model}. The encoder is custom code that runs once at the beginning to turn the structured data into tensors. The RNN model itself consists of the part in the dashed box. It generates one output symbol per time step and stops when it generates a stop symbol.

\subsection{Structured data encoding}
The structured data is a set of named argument values, where every value is either a string, a number, a Boolean, or an enum. We assume that the NLG application has a fixed schema of possible arguments, and a fixed schema of possible enum values. Each example can use a different set arguments, and the arguments normally vary based on the domain and intent of the example.

We encode strings as symbols using a sentence piece model, and we refer to each symbol, argument, or enum value by its index in the sentence piece vocabulary or schema. We treat numbers as strings and Boolean values as enums. Using this mapping, we conceptually encode the structured data as a table where we lay out the argument values vertically, with one string symbol or enum value per row. For example:

\newcommand{\eol}{{\textless}eol{\textgreater}}

\begin{description}
\item[Structured data] \hfill \\
   num\_tickets: 4 \\
   theatre: "Century 16" \\
   time: "8:00 pm" \\
   domain: MOVIES \\
   intent: NOTIFY\_SUCCESS \\
   num\_slots: 4
\item [Data table] \hfill \\\\
\begin{tabular}{llll|c|l}
     56  &  4  &  1  &  0  &  4  & num\_tickets  \\
     14  &  4  &  1  &  1  &  \eol\\
     71  &  5  &  9  &  0  &  C  & theatre \\
     105 &  5  &  9  &  1  &  e \\
     114 &  5  &  9  &  2  &  n \\
     120 &  5  &  9  &  3  &  t \\
     121 &  5  &  9  &  4  &  u \\
     118 &  5  &  9  &  5  &  r \\
\end{tabular}
\begin{tabular}{llll|c|l}
     125 &  5  &  9  &  6  &  y \\
     36  &  5  &  9  &  7  &  \textvisiblespace \\
     53  &  5  &  9  &  8  &  1 \\
     58  &  5  &  9  &  9  &  6 \\
     14  &  5  &  9  &  10 &  \eol \\
     60  &  7  &  9  &  0  &  8 & time \\
     62  &  7  &  9  &  1  &  : \\
     52  &  7  &  9  &  2  &  0 \\
     52  &  7  &  9  &  3  &  0 \\
     36  &  7  &  9  &  4  &   \textvisiblespace \\
     116 &  7  &  9  &  5  &  p \\
     113 &  7  &  9  &  6  &  m \\
     14  &  7  &  9  &  7  &  \eol \\
     5   &  30 &  8  &  0  &    & domain \\
     11  &  31 &  8  &  0  &    & intent \\
     56  &  32 &  1  &  0  &  4 & num\_slots \\
     14  &  32 &  1  &  1  &  \eol \\
\end{tabular}
\end{description}

\noindent The data table is the four columns of numbers---the two columns to the right are simply to show the symbol and argument that each row encodes. Although the table shows one character per row for illustration, the sentence piece vocabulary groups these into larger chunks, and each row is usually a multi-character sentence piece. The columns in the data table are:

\begin{figure*}
\includegraphics*[page=8,width=6in,trim=0.5in 1in 0.5in 01in]{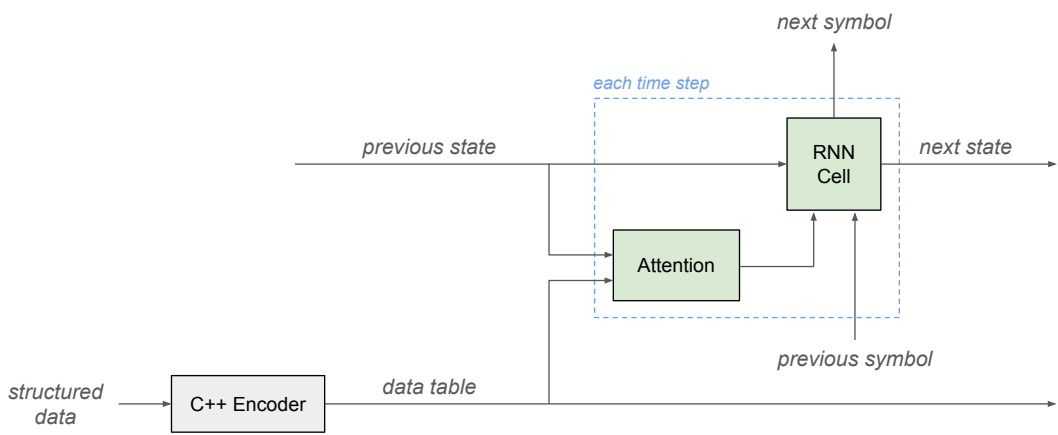}
\caption{Architecture of the distilled data-to-text RNN model}
\label{data-to-text-model}
\end{figure*}

\begin{enumerate}
    \item The symbol index or enum index
    \item The argument index in the schema
    \item The argument type index
    \item The position of the symbol in the argument
\end{enumerate}

\noindent The data table does not include rows for the argument names, since this information is already implicitly present in the argument index in column 2. We have experimented with additional annotation columns, for example including the position of a feature or symbol in the example as a whole. In our application the order of the structured data arguments is irrelevant, though, and the four columns shown capture this basic invariance property and work well.

To finish the encoding, we expand each element of the table using an embedding matrix of width $E$, and then project the resulting tensor of width $4E$ to smaller key and value tensors of width $W_k$ and $W_v$. The final tensors have shape $[B, S, W_k]$ and $[B, S, W_v]$, where $B$ is the batch size and $S$ is the maximum number of symbol and enum rows in an encoded example. We typically use $B=256$ and $E=W_k=W_v=64$. The value of $S$ depends on how many features are in a typical example in the NLG task, and how long the encoded feature values are. In our application $S$ is typically around 400 using a vocabulary of 2048 sentence pieces.

\subsection{RNN cell and attention mechanism}
For the RNN cell we normally use a single layer, layer normalized LSTM cell with 1024 hidden units. At every time step, the model reads from the structured data by projecting the previous RNN state to a query tensor of shape $[B, W_k]$. We use scaled dot product attention to compare the query against the key for each row, then return a weighted sum of the values. The output has shape $[B, W_v]$. We concatenate this with the embedding of the previous output symbol and feed the result to the RNN cell. To compute the next symbol we use a softmax projection of the RNN output to the sentence piece vocabulary.

We have done extensive experiments with variations of this model, including multiple read heads, multiple layers, and self attention to the previous RNN states. These are all capable of producing gains depending on the application. However, for our main task the simple single layer model with a single read head and no self attention works well.

We add a small number of delay steps at the beginning of each example so that the RNN cell has a chance to read from the data table before it has to start producing output. This is important in case the model needs to read multiple values before deciding on the first symbol. For example, if the desired output is "Yes, the Cubs won 4-2", then the model may need multiple steps to figure out the domain, the intent, and whether the correct answer starts with "Yes" or "No". We treat the number of delay steps as a hyper-parameter and empirically set it to 3 for our task. Using multiple read heads together with a deeper model can eliminate the need for these delay steps.

It is interesting to compare our data-to-text model to a more modern Transformer model. Our model predates Transformer, and we have not switched as the  RNN provides a good balance of inference speed and quality for our task. One difference compared to Transformer is that our model is effectively decoder only: there is no learned encoding other than the embedding matrix. At the same time, the columns in the data table are similar to what a Transformer encoder might produce from a flat input representation of the argument names and values. Also, even though there is no learned encoder, the model can do additional NLU over time using the RNN state in parallel with generating output. A second difference is self attention to previous states, and when we add this to our model, the architecture is very similar to a decoder only Transformer. While our model is competitive on the NLG tasks that we have tried it, we do see a gap compared to Transformer on the public WebNLG data set \cite{gardent2017webnlg}. At the end of the paper we talk about how we see the data-to-text model evolving and converging with the NMT model, where Transformer will be the appropriate starting point.

\subsection{Model training}
We train the model in our NLG task for 1M steps, which is about 5 epochs at our usual data sizes.  We use a batch size of 256, 10\% dropout, a learning rate of 6e-3, and a learning rate schedule that uses a short linear warm up followed by exponential decay.

\subsection{Training data}
\label{synthetic-examples}
We use synthetic examples for training to preserve user privacy, and because we do not always have access to real examples from logs at scale. By using synthetic examples we can create as much training data as we need. Our synthetic example generator uses knowledge about the argument schema to pick random argument values for each example. This requires some care to get meaningful argument values that are close enough to the inference distribution that the final model works well.

For example, suppose our NLG task allows a user to set a timer, generating sentences like this one:

\begin{quote}
    Okay, setting a timer for 1 hour, 5 minutes, and 37 seconds.
\end{quote}

\noindent Also suppose that the input argument is a single time value denominated in seconds. If we randomly pick training values between 1 and 100,000, the training data will be badly out of sync with real requests that are almost all even multiples of minutes and hours:

\begin{quote}
    Okay, setting a timer for 5 minutes.
\end{quote}

\noindent The model may perform perfectly on training examples like the first sentence, but poorly on real requests like the second one that occur less than 1 time in 1000 in the training data.

To avoid this problem, we use schema annotations to provide semantic information about each argument. The synthetic example generator uses these annotations to create meaningful values. For example, we can annotate an argument as a user supplied time value, in which case the example generator will know to over sample even multiples of hours, minutes, and days. As shown in Figure~\ref{example-generator}, the example generator supports annotations for common data types like dates, times, and numbers, as well as letting us specify enum types for enum arguments and entity types from the Google Knowledge Graph \cite{singhal2012KG} for entities.

\begin{figure}[t]
\includegraphics*[page=10,width=3in,trim=2in 1in 2in 0.75in]{figures.pdf}
\caption{Synthetic example generation}
\label{example-generator}
\end{figure}

We run all of the synthetic training examples through the English NLG system and the NMT models to create training pairs that go from structured data to localized text in each of our target languages. We use these to train our distilled data-to-text models. We run the English NLG system in a mode that lets us produce all of the possible English variants, and we use all of these and their NMT translations as reference texts.

We train an English model in addition to the non-English models. This is actually an important starting point, as it lets us test the quality of the data-to-text model in English to ensure that the structured data contains enough information for the model to do its basic task, independent of translation. As a practical matter, it's also much easier for us to first identify and fix problems in English, as we do not always speak the languages that we are working in.

\begin{figure*}
\includegraphics*[page=9,width=6in,trim=0.75in 2.25in 1in 2in]{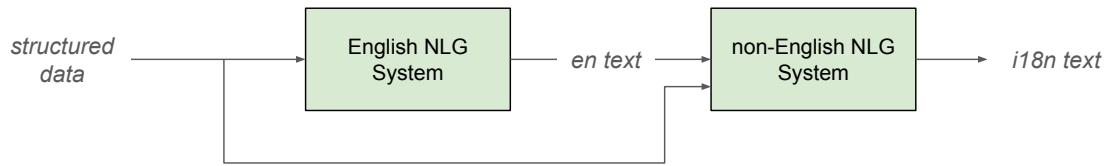}
\caption{Inference architecture}
\label{prod-system}
\end{figure*}

We have found from the English experiments that it's important in our particular NLG task to augment the original structured data with additional features. For example, if the English text contains the word "Thursday", then "Thursday" must appear somewhere in the structured data, since otherwise the data-to-text model needs to figure out how to convert a raw date to the corresponding day of the week---a hard problem. In our task the English NLG system is able to mark potentially useful spans of text, and so the simplest solution is to include all of these as additional structured data arguments. We have also experimented with using the schema information to add additional features, for example adding a day of the week feature for each date. This also works, though we have found that using spans of text in the English output gives better coverage and can take advantage of very specific domain and intent logic buried deep in the English system.

In fact, we have found that using the spans of English text can eliminate the need for almost all of the original structured data. This is not surprising, as the English text already contains almost all of the relevant information in the structured data, and it often contains more as it reflects knowledge from other sources that is in the English NLG system, for example figuring out the day of the week for a date. One place where we do still need the structured data is to get localized text for entities and strings, as without this the data-to-text model needs to either memorize translations or learn how to transliterate. We'll have more to say about localizing entities in the final section on future work.

\subsection{Evaluation}
We use both synthetic and real examples for evaluation. The synthetic examples provide better coverage of domains and intents, while the real examples provide more realistic argument values and ensure that the distilled model works well on the actual distribution of examples that we expect to see at inference time.

To evaluate the English model we use exact match against all of the English reference texts. The accuracy is very close to 100\% on both the synthetic and real test sets, though occasionally we find examples where the model uses a slightly different variation, for example different punctuation. Based on hand inspection these variations are almost never wrong and are often better than the English NLG system, revealing bugs and inconsistencies. Where there are errors, we can trace them to problems in the training data and fix them by fixing these problems.

To evaluate the non-English models we use two automatic metrics: exact match against all of the NMT translated English references, and the output of the accuracy error detection model that we created to filter the NMT data. These metrics are both over 99\% on our synthetic eval sets and in the 95-97\% range on our real eval sets. From looking at the examples, the gap to 100\% is a combination of underlying variation in the NMT output, errors in the accuracy error detection model, differences in the inference and training distributions, and a small but definite gap in the quality of the distilled data-to-text output as compared to the original NMT models. The data-to-text model fulfills our purpose, though, as it reaches the quality level and performance that we are targeting for our application.

\section{Practical Considerations}
Incorporating our localization work in a working system requires many additional engineering pieces. For our particular NLG task this includes a training pipeline to continuously train the NMT and data-to-text models at scale, a dashboard to monitor quality, tools to send examples for human evaluation and translation, integration with serving systems to use the models on real requests, and processes to keep up with changes to the English NLG output. We share what we have learned about keeping up with changes to the English output, as this seems generally applicable to localization in NLG tasks.

Figure~\ref{prod-system} shows the schematic flow of how our NLG application performs inference. We first use the English NLG system to generate English, and then use the English and the structured data to produce the localized output.

The English and non-English NLG systems can potentially change at different times. Changing the English system without changing the non-English system can cause problems, because the localized data-to-text models may then see examples that are different than the training distribution. There are two kinds of problematic changes:

\begin{description}
    \item[Changes to the structured data schema] \hfill \\
    This usually happens because we add a new domain or intent, but it can also happen if we refactor an existing intent, for example adding or deleting an argument.
    \item[Changes to the English output] \hfill \\
    Our distilled data-to-text model uses spans of text in the English output as features, so changing these affects the model.
\end{description}

Our solution is to deploy two versions of the English system: one for English requests, and one for non-English requests. The one that we use for non-English requests is always identical to the one that we use to train the NMT and distilled data-to-text models. We have designed the system so that we can incorporate and test changes to the English system independently, so that changes can flow through to the non-English system at different rates. For example, we may be able to validate that small changes to the English output work fine with the existing models, so we can incorporate these changes immediately, while we need to use a longer process when we add a new domain so that we can collect new translations. Our process allows us to make these changes independently. Figuring out how to safely make changes like this has been an important aspect of letting us build a localized NLG system at scale.

\section{Future Work}
\subsection{Unified model}
Our current work focuses on combining the NMT and data-to-text models into a single model. This model goes from annotated English text and structured data to localized text, where the annotations and structured data give the model additional context that is not present in the text. The annotations are for information that maps to a span, for example marking a span as an entity, or indicating that 1/2 is a fraction and not a date. The structured data is for information that is not present in the English text, for example the gender of the speaker, or whether the output should be formal or colloquial, or the desired output length. To create training data at scale, we extend the NER entity tagging to annotate other useful spans of text in our NMT data, and we use text classification to add labels like GENDER=M and STYLE=FORMAL. This converts our NMT training corpus into one that contains structured data and annotations that we expect to be able to draw on at inference time.

\subsection{Transliteration and copying}
One other area of current work is improving how we transliterate and copy names. Sometimes we want to translate a name into a local variant, for example translating "France" to "Frankreich" in German, but most of the time we want to copy or transliterate, where the choice depends on whether the target language uses the same alphabet as the source. To improve quality, we would like to take advantage of curated translations and transliterations, for example from the Google Knowledge Graph. We also want to correctly copy spans of text in the English input that are already in the target language. This often happens in NLG tasks where the output includes user generated content.

To solve these problems, we extend our entity tagging to include the proposed localized text in the input, and we use placeholders anywhere that the target text copies the input. For example:

\begin{description}
\item[English] \hfill \\
    Tomorrow's game is in \textcolor{blue}{{\textless}location, Frankreich{\textgreater}}France\textcolor{blue}{{\textless}/location{\textgreater}}.
\item[Target text] \hfill \\
    Das Spiel morgen ist in \textcolor{blue}{\$0}.
\end{description}

\noindent We add the \$0 with a preprocessing step that recognizes that "Frankreich" appears verbatim in the output. The model learns to generate the \$0 placeholder, which we then replace during postprocessing. This technique has the nice property of replacing spans of text with placeholders wherever possible, while relying on the model to generate the full text whenever it's not possible to copy text verbatim. In either case, the localized name in the input is a significant help to inflect the final sentence correctly and use a proper local name.

\bibliographystyle{acl_natbib}
\bibliography{paper.bib}

\end{document}